\documentclass[11pt]{article}

\usepackage[final]{acl}

\usepackage{times}
\usepackage{latexsym}

\usepackage[T1]{fontenc}

\usepackage[utf8]{inputenc}

\usepackage{microtype}

\usepackage{inconsolata}

\usepackage{graphicx}
\usepackage{algorithm}
\usepackage{algorithmic}
\usepackage{amsmath}
\usepackage{booktabs} 
\usepackage{multirow}
\usepackage{amsfonts}
\usepackage{bm}

\title{Detecting Hallucinations in Retrieval-Augmented Generation via Semantic-level Internal Reasoning Graph}

\author{
	Jianpeng Hu \textsuperscript{1}, 
	Yanzeng Li \textsuperscript{2}, 
	Jialun Zhong \textsuperscript{1}, 
	Wenfa Qi \textsuperscript{1}, 
	Lei Zou \textsuperscript{1}\thanks{~~Corresponding author} \\
	\textsuperscript{1} Wangxuan Institute of Computer Technology, Peking University, Beijing, China \\
	\textsuperscript{2} Institute of Artificial Intelligence and Future Networks, Beijing Normal University, Zhuhai, China \\
	\texttt{jianpeng.hu@outlook.com, liyanzeng@bnu.edu.cn, zhongjl@stu.pku.edu.cn}\\
	\texttt{\{qiwenfa, zoulei\}@pku.edu.cn}}

\begin{document}
	\maketitle
	\begin{abstract}
		The Retrieval-augmented generation (RAG) system based on Large language model (LLM) has made significant progress. 
		It can effectively reduce factuality hallucinations, but faithfulness hallucinations still exist. 
		Previous methods for detecting faithfulness hallucinations either neglect to capture the models' internal reasoning processes or handle those features coarsely, making it difficult for discriminators to learn. 
		This paper proposes a semantic-level internal reasoning graph-based method for detecting faithfulness hallucination. 
		Specifically, we first extend the layer-wise relevance propagation algorithm from the token level to the semantic level, constructing an internal reasoning graph based on attribution vectors. 
		This provides a more faithful semantic-level representation of dependency. 
		Furthermore, we design a general framework based on a small pre-trained language model to utilize the dependencies in LLM’s reasoning for training and hallucination detection, which can dynamically adjust the pass rate of correct samples through a threshold.
		Experimental results demonstrate that our method achieves better overall performance compared to state-of-the-art baselines on RAGTruth and Dolly-15k. 
	\end{abstract}

	\section{Introduction}	
	LLMs easily generate grammatically coherent but factually incorrect outputs, a phenomenon commonly referred to as ``hallucination''~\cite{mishrafine, zhang2024enhancing, li2023inference}. 
	Post-learning for downstream tasks or introducing the RAG system~\cite{lewis2020retrieval} can mitigate the factuality hallucinations to some extent, which refer to the tendency of LLMs to produce outputs that are inconsistent with real-world facts. 
	However, due to the inherent knowledge bias of the internal parameters of LLMs, the generated content of RAG may be inconsistent with the context provided by users~\cite{Towards_Faithful}, leading to faithfulness hallucinations.
	\begin{figure}[t]
		\centering
		\includegraphics[height=0.24\textwidth,width=0.5\textwidth]{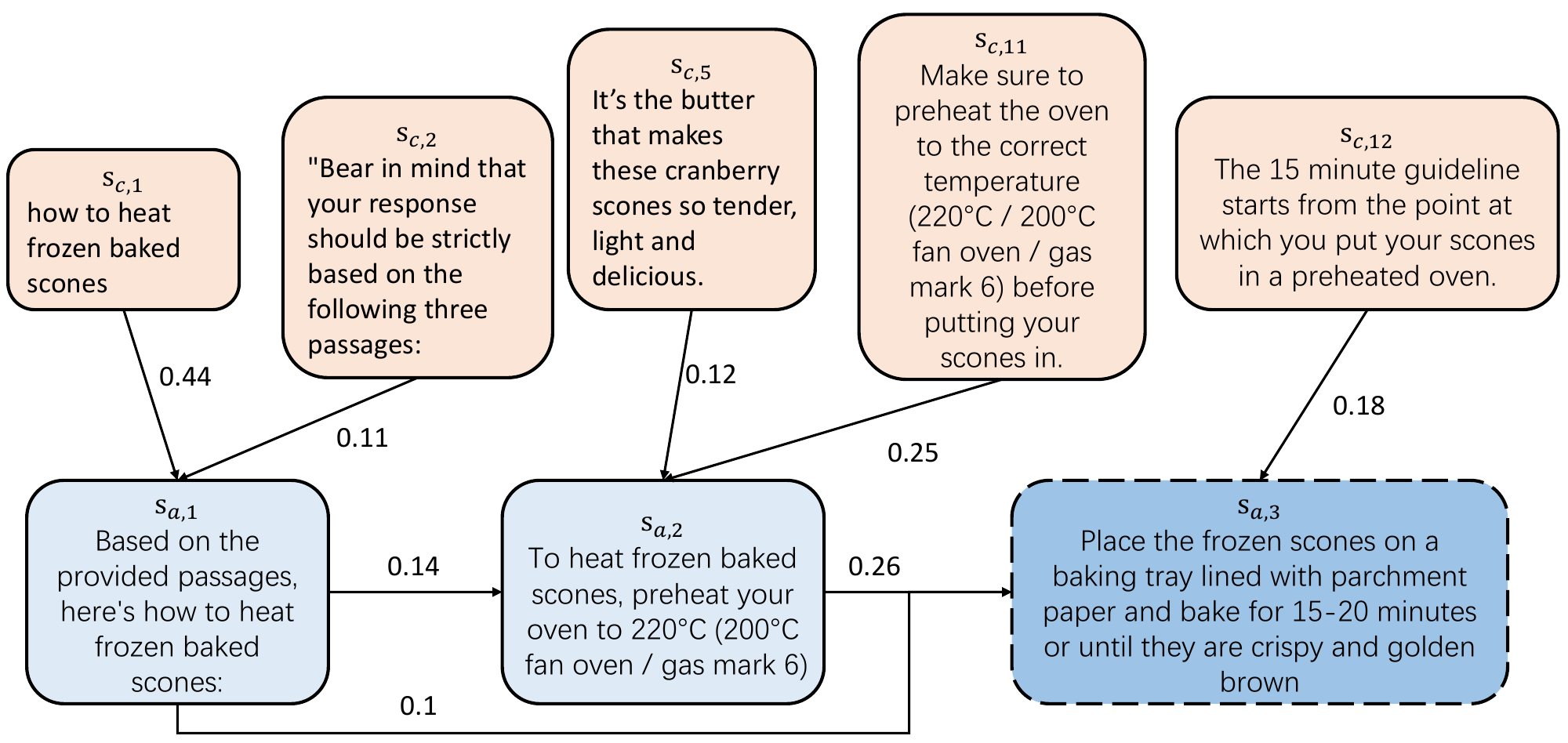}
		\caption{Example of a semantic-level internal reasoning graph. Yellow nodes represent contextual semantic fragments, blue nodes represent semantic fragments of the model's response, and the weight on the edge indicates the contribution degree of the source semantic fragment to the target (with an upper boundary of 1). The dashed box indicates a hallucinated semantic fragment.}
		\label{graph}
	\end{figure}
	\citet{he2022rethinking} has found that faithfulness hallucinations arise primarily from the inconsistency between LLMs' word-level output and true thought process, meaning that LLMs only utilize surface knowledge, such as entity popularity~\cite{Knowing_the_Facts}, during reasoning. 
	
	To detect faithfulness hallucinations, both~\citet{SelfCheckGPT} and~\citet{HalluClean} propose post-processing methods with LLMs. However, multiple invocations of the LLM system can lead to significant resource consumption and amplification of model bias.
	Other scholars~\cite{Do_LLMs_Signal, Synchronous_Faithfulness, DISCOVERING_LATENT} assess hallucinations based on the internal embedding of the model, which often relies on heuristic discrimination trained on these abstract features, resulting in poor interpretability. 
	\citet{LRP4RAG, Lookback_Lens} detect hallucinations from the perspective of output attribution, but their direct accumulation of all token-level attribution vectors for LLMs' responses will introduce substantial noise (as shown in Appendix \ref{sec:noise}).
	
	To better explain the origin of faithfulness hallucinations, motivated by \citet{Peering_into}, we divide the tokens generated during the autoregressive reasoning of LLMs into linking and substantive tokens. 
	Unlike the original concept, we define linking tokens as non-substantive text in LLM-generated responses that serve to connect contextual information, whereas substantive tokens refer to text that utilizes the contextual information provided by users and reflects the semantic content of the responses.
	We consider that \textbf{hallucinations originate from LLM mistakenly generating substantive tokens as linking}, which manifests on the surface as generation based on entity popularity.
	Visual analysis from the perspective of attribution score distribution can reveal which substantive tokens LLMs treat as linking tokens, and these differences are difficult to detect from a human perspective because of semantic drift.
	Specifically, linking tokens rely more on words generated earlier within the same sentence, whereas substantive tokens also depend on words in the long-distance context (detailed in Appendix \ref{sec:glue}).
	
	Since faithfulness hallucinations typically occur at the semantic level, we extend the aforementioned concepts to this level and detect faithfulness hallucinations by our proposed semantic-level internal inference graphs, which can faithfully construct the dependency of the linking and substantive fragments.
	Specifically, we first employ Layer-wise relevance propagation (LRP) to calculate the score vector attributed to each token during the autoregressive process. 
	This attribution method, which utilizes internal model parameters and predefined rules, faithfully reflects the true computational process within the model. 
	According to the token-level attribution vectors, we model the attribution relationship between contextual semantic fragments and those of LLM's response, forming the internal reasoning graph, as illustrated in Fig. \ref{graph}.
	We can observe that the semantic fragments with hallucinations assign higher attribution scores to previously generated semantic fragments in the model's autoregressive reasoning, while showing weaker dependency on the context provided by the user. 
	LLM mistakenly treats substantive fragments as linking during the reasoning process, leading to the hallucination phenomenon. 
	
	Based on the above observation, we linearize each response node and attribution dependency in the internal reasoning graph, concatenate them into a prompt, and input it to downstream pretrained language models (PLM) for fine-tuning in binary classification tasks. 
	In the hallucination detection phase, we determine hallucination based on the binary classification label distribution of all semantic fragments in the LLM-generated text. 
	The entire process relies solely on the model with a small number of parameters.
	
	In summary, our contributions mainly include:
	\begin{itemize}
		\item We extend token-level LRP to the semantic level and propose a method for constructing semantic-level internal reasoning graph of LLMs.
		\item We analyze faithfulness hallucinations from the distribution differences between linking and substantive fragments and utilize internal reasoning graphs to detect them.
		\item The experimental results on two general datasets demonstrate that the performance of our framework surpasses previous baselines.
	\end{itemize}	
	
	\section{Related Work}
	\paragraph{Hallucination Detection}
	LLMs often produce content that is grammatically correct but semantically conflicts with real-world or contextual knowledge, known as hallucinations~\cite{Survey_Hallucination}. Hallucinations widely exist in downstream reasoning~\cite{FreshLLMs, Dawn_After_Dark}. Existing hallucination detection methods mainly include: (1) LLMs-based post-verification methods, which detect hallucinations by using multi-agent systems~\cite{Hallucinate_Less_Thinking}, multiple rounds of self-criticism~\cite{Self-contradictory_Hallucinations}, or consistency of multiple generations~\cite{SEMANTIC_UNCERTAINTY}; (2) representation-based methods, which identify abnormal states through model hidden states~\cite{Distributional_Semantics_Tracing}, outputs of attention modules~\cite{Constructing_benchmarks}, semantic alignment rates~\cite{Zero-shot_Faithful}, etc.; (3) task-specific methods, including fine-tuning~\cite{HalluGuard} or designing task-specific features~\cite{multilingual_translation}. Most of the aforementioned methods rely on explicit reasoning or uninterpretable feature spaces, which may fail to detect semantic biases within the model. Our framework focuses more on faithfully modeling the true reasoning dependencies within LLMs and training a lightweight detector based on them.
	\paragraph{Faithful Attribution}
	Our method, grounded in additive interpretability theory~\cite{Neural_additive}, decomposes model predictions into a sum of contributions from each input, analyzing the significance of source tokens to target tokens based on contribution scores. Perturbation-based method is the most classic attribution algorithm, including SHAP~\cite{unified_approach}, LIME~\cite{why_should_trust}, AtMan~\cite{Atman}, etc. These methods exhibit significant computational complexity in the application of LLMs. For the transformer architecture, some scholars utilize attention mechanisms~\cite{attention_flow, attention_visualization} to capture causal relationships. However, these methods lack category specificity and cannot faithfully interpret the final predictions. 
	Based on an improved backpropagation method, the propagation rules are customized between layers to trace back from the model output to the input. Classic methods include Input $\times$ Gradient~\cite{inside_convolutional}, LRP~\cite{pixel-wise, Analyzing_source}, SmoothGrad~\cite{Smoothgrad}, etc.  AttenLRP~\cite{AttnLRP} employed in this paper is an improved backpropagation method that can effectively handle nonlinear relationships.
	
	\section{Methodology}
	\subsection{Task Formulation}
\begin{figure*}[t]
		\centering
		\includegraphics[height=0.542\textwidth,width=1\textwidth]{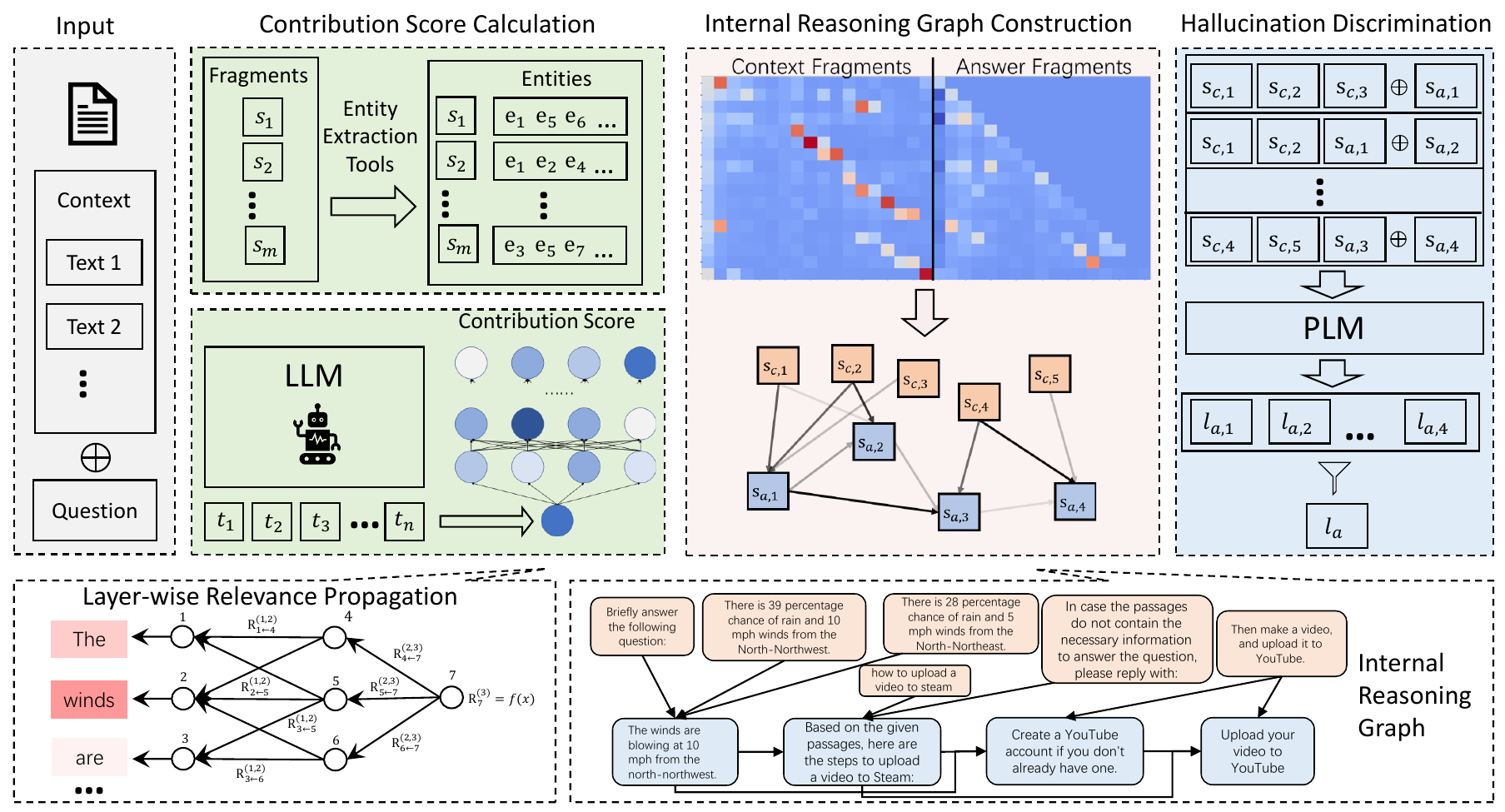}
		\caption{The framework of our method. LRP is first applied to derive a relevance distribution $R_i$ based on the parameters of the LLM. Based on $R_i$ of entities and semantic fragments of input and output content, a semantic-level internal reasoning graph of LLMs is constructed. Subsequently, a PLM is used to determine whether each fragment exhibits semantic conflicts or omissions. Finally, the degree of hallucination occurring in the reasoning graph units is used to determine whether the model's overall response exhibits hallucinations.}
		\label{framework}
	\end{figure*}

	Let \(D=(Q_i,A_i)_{i=1}^{|D|}\) denote a RAG dataset consisting of \(|D|\) samples, where each sample pair comprises a user query \(Q_i\) and its corresponding answer \(A_i\). Each answer is based on contexts retrieved from a knowledge base \(C=\{C_j\}_{j=1}^{|C|}\), where \(C_j\) represents a text block. Given a question \(Q_i\), a retrieval model parameterized by $\phi$ first retrieves the most relevant text blocks from \(C\) to form a prompt, which is then input into a generative model parameterized by $\theta$ to generate the answer $A_i$. The complete process of RAG is defined as follows:
	\begin{equation}
		\begin{split}
			& P(A_i | Q_i)=P_{\phi} (C_j | Q_i) P_{\theta} (A_i | Q_i,C_j)   \\
			& P_{\theta} (A_i | Q_i,C_j )=\prod_{k=1}^n P_{\theta} (a_k | a_1, \cdots ,a_{k-1};C_j;Q_i)
		\end{split}
	\end{equation}
	where $n$ is the total number of tokens in the target answer $A_i$. We introduce a discriminator parameterized by $\gamma$ to determine whether $A_i$ contains hallucinations:
	\begin{equation}
		P(L_i|Q_i,C_j)=P_{\theta} (A_i | Q_i,C_j) P_{\gamma} (L_i |Q_i,C_j,A_i)
	\end{equation}
	where label $L_i$ indicates whether the sample exhibits hallucinations.
	The framework of our method is shown in Fig. \ref{framework}.  
	
	\subsection{Contribution Score Calculation}
	The LRP algorithm is used to calculate the contribution score. 
	The basic assumption of LRP is that a function \(f_j\) with \(N\) input features \( \bm{x}=\{ x_i \}_{i-1}^N \) can be decomposed into independent contributions of a single input variable \( R_{i \leftarrow j}\), representing the amount of output \(j\) that can be attributed to input \(i\). When these contributions are added up, they are proportional to the original function value. 
	\begin{equation}
		f_j (\bm{x}) \propto R_j = \sum_i^N R_{i \leftarrow j}
	\end{equation}
	The decomposition property of LRP leads to an important conservation property~\cite{AttnLRP}, which ensures that the sum of all contribution scores in each layer remains constant. 
	This feature allows for meaningful and faithful attribution, as the scale of each contribution score can be associated with the output of the original function. 
	
	In the generation stage, when generating each token \(a_k\), the generator simultaneously outputs a correlation vector \( \vec{r}_k\) based on internal gradients and predefined LRP rules (see Appendix \ref{sec:lrp} for details), which is used to quantify the correlation between \(a_k\) and each contextual token \(C_j\). 
	By aggregating all correlation scores \( \vec{r}_k\) of a token during the generation process, we can obtain a correlation matrix \(R_i\).
	
	\subsection{Internal Reasoning Graph Construction}  
	Due to potential semantic biases in the model during training, which is caused by the solidification of pre-trained corpora and the prejudice of the model, the semantic meaning of certain words understood by LLMs may differ from their meaning in the real world. 
	This results in a gap between the reasoning process understood by humans based on chain-of-thought and the reasoning process actually intended by LLMs.
	That is, there are some thinking outputs that cater to human preferences, while the model's true reasoning process does not rely on these thinking outputs or follow thinking shortcuts.
	The correlation matrix $R_i$ calculated in the previous step using the LRP is a token-level attribution matrix that is faithful to the internal reasoning process of the model. 
	Therefore, the internal reasoning graph constructed based on $R_i$ can faithfully reflect which contextual semantic fragments a certain semantic fragment originates from during the internal inference of the model.
	
	We first recursively use ``\textbackslash n'' and sentence tool of Spacy to segment the input context into individual semantic fragments, thereby obtaining a set $S_c = \{s_{c,1}, s_{c,2}, \cdots \}$, where each $s_{c,i}$ represents the $i$th semantic fragment in that context. 
	For the model's output, we similarly use ``\textbackslash n'' to segment, obtaining a set of semantic fragments $S_a = \{s_{a,1}, s_{a,2}, \cdots \}$ contained in one response from the model. 
	The union of these semantic fragments, denoted as $S = S_c \cup S_a$, serves as the node set of the model's internal reasoning graph, that is, the set of atomic steps for model inference.
	
	As mentioned above, the text within a semantic fragment includes both linking content used to connect contexts and make them grammatically correct, and substantive content that specifically reflects the actual meaning and thought expressed in the text. 
	When calculating attribution at the fragment level based on LRP, the attribution scores calculated for these linking fragments introduce a significant amount of noise. 
	To distinguish between semantic fragments that contain rich semantic information and meaningless content, we use a general named entity extraction tool (Spacy and Stanze) to extract entities from the document to the greatest extent possible.
	The extracted content includes nouns, verbs, noun phrases, negation words, and named entities, which can maximize the reflection of the semantic meaning expressed in the text while filtering out linking content. 
	The set of extracted results, after removing duplicates, is considered the core content $E = \{e_1, e_2, \cdots \}$ that expresses semantic information in the text. 
	Finally, by mapping the set $E$ to each element in the semantic fragment set $S$, we can obtain the subset of actual meaning $E_{s_i} = \{e\}$ contained in each semantic fragment.
	
	In each target semantic fragment $s_i$, we only select tokens contained in $E_{s_i}$ to calculate its attribution vector relative to the preceding text. 
	Then, we average all the selected correlation vectors element by element to obtain the token-level attribution vector of the semantic segment relative to the preceding text. 
	For the attributed semantic fragment $s_j$ in the preceding text, we select the maximum value of tokens contained in $E_{s_j}$ in the attribution vector of $s_i$ as the score of $s_i$'s attribution to $s_j$. 
	When attributing, $s_i$ often only attaches to a small number of tokens with actual meaning in the preceding text. 
	Using an average function can dilute the high correlation information in $s_j$ due to fragment length. 
	Therefore, we use a maximum function to aggregate the correlation vectors:
	\begin{equation}
		W_{ s_{a,i},s_{*,j} }=\left \{
		\begin{aligned}
			\underset{w \in s_{c,j}} {max} & (\frac{1}{|E_{s_{a,i}}|} \sum_{e \in E_{s_{a,i}} } \vec r_{e, w} ) \\ 
			&\quad s_{a, i} \in S_a, s_{c, j} \in S_c\\
			\underset{w \in s_{a,j}} {max} & (\frac{1}{|E_{s_{a,i}}|} \sum_{e \in E_{s_{a,i}} } \vec r_{e, w} ) \\ 
			&\quad s_{a, i},s_{a, j}  \in S_a, i > j\\
			0 &\quad s_{a, i},s_{a, j}  \in S_a, i \le j \\
		\end{aligned}
		\right
		.
	\end{equation}
	where $\vec{r}_{e,w}$ represents the element associated with token $w$ in the vector corresponding to entity $e$ in the matrix $R_i$. If the entity consists of multiple tokens, the average of the vectors corresponding to these multiple entities is taken. 
	Through this step, we can obtain the semantic-level correlation matrix $W \in \mathbb{R}^{n_a \times (n_c + n_a)}$, where each element $W_{i,j}$ represents the influence degree of the $j$th semantic fragment on the $i$th. As shown in Fig. \ref{framework}, since subsequent semantic segments do not have an attributive influence on preceding semantic segments, this matrix is a lower triangular matrix from column $n_a$ to column $n_c + n_a$.
	
	The internal reasoning graph $G = \{V, E\}$ is a directed graph that faithfully reflects the dependency relationships between semantic fragments during the internal inference process of the model. 
	The nodes of the graph, $V = S_c \cup S_a$, represent the semantic fragments obtained from the previous context. 
	The edges are associated through the attribution scores between semantic fragments. The weight of each edge is the normalized attribution score. Based on the semantic-level correlation matrix, we propose two methods for constructing intra-inference graphs:
	\paragraph{Top k method}
	This strategy first ranks the attribution scores calculated for the target semantic fragments from highest to lowest, then selects the top $k$ fragments as the source, and inserts edges from the source to the target semantic fragments in the graph. 
	The set of incoming edges for node $s_{a,j}$ can be represented as:
	\begin{equation}
		E_{a,j}=\{(s_{*,i},s_{a,j})|Topk(W_{s_{a,j},s_{*,i} }),s_{*,i} \in V\}
	\end{equation}
	\paragraph{Adaptive Method}
	The distribution of attribution scores calculated often exhibits a long-tail characteristic. 
	To select edges adaptively, we arrange the attribution scores in descending order and calculate the discrete gradient of the sequence. 
	The maximum discrete gradient point is used to distinguish between important and unimportant source semantic fragments. 
	Assuming that $v_1 \ge v_2 \ge \cdots \ge v_{n_c+n_a}$ is the non-increasing ordering of elements in $W_{s_{a,i}}$, where each $v_1$ is mapped one-to-one with $s_{*,i}$ using the function $f(\cdot)$, then the set of incoming edges for node $s_{a,j}$ can be represented as follows:
	\begin{equation}
		\begin{split}
			& m=\arg \underset{1 \le i \le n_c+n_a-1} {max} (W_{s_{a,j},f(v_i)}-W_{s_{a,j},f(v_{i+1})})\\
			& E_{a,j}=\{(s_{*,i},s_{a,j})|s_{*,i}\in f(v_k),1\le k \le m \}
			\label{eq: grad}
		\end{split}
	\end{equation}
	
	The union of the incident edge sets of all nodes forms the edge set $E = \{E_{a,j} | 1 \le j \le n_a\}$.
	
	\subsection{Hallucination Discrimination}
	As depicted in Fig. \ref{graph}, in the attribution nodes of hallucinated semantic fragments, a higher proportion of attribution is allocated to the previous semantic fragment answered by the model; whereas the attribution nodes of non-hallucinated semantic fragments tend to be more related to the contextual corpus provided by the user, indicating that they are more faithful to the context provided by humans. 
	Therefore, an important reason for the occurrence of hallucination is that the model treats the next generated semantic fragment as linking content, rather than as substantive content.
	
	To enable the model to discover the attribution distribution differences and semantic differences in these contextual dependencies, we linearize the inference graph into multiple semantic combinations.
	Specifically, for each semantic fragment $s_{a,j}$ answered by the model, we concatenate all its incoming edges to form a prompt, and feed it into a pre-trained language model (PLM) to obtain the label $l_{a,j}$:
	\begin{equation}
		l_{a,j}=PLM(\{s_{*,i} | s_{*,i}\in E_{a,j} \} \oplus s_{a,j})
	\end{equation}
	
	In this paper, we utilize the ALIGNSCORE~\cite{ALIGNSCORE}, which is based on the RoBERTa architecture~\cite{liu2019roberta}, as the PLM for training, and employ its binary classification inference head for hallucination discrimination. 
	ALIGNSCORE has undergone pretraining on a vast amount of data, focusing on the degree of information alignment between two arbitrary segments. 
	Therefore, by simply fine-tuning it with the downstream cross-entropy loss function, it can exhibit strong hallucination detection capabilities.
	Ultimately, we employ a flexible threshold $\alpha$ to determine whether the entire model response exhibits hallucinations.
	If the proportion of semantic fragments containing hallucinations in the model response exceeds $\alpha$, then the model response is considered to exhibit hallucinations. Formally, it can be expressed as follows:
	\begin{equation}
		l_a = \mathbb{I}[ \frac{\sum_{j=1}^{n_a} \mathbb{I}[l_{a,j}=0]}{n_a} \le \alpha ]
		\label{eq:class}
	\end{equation}
	where $\mathbb{I}$ denotes an indicator function. This paper aligns with previous research, where $l_a=1$ signifies a correct model response, indicating the absence of hallucination, and $l_a=0$ denotes an incorrect model response, indicating the presence of hallucination. 
	When $\alpha=0$, it means that any semantic fragment in the model response that is suspected of being hallucinatory will result in the entire response being classified as hallucination. 
	In practice, the value of $\alpha$ can be adjusted based on the required level of model reliability for specific scenarios.
	
	\section{Experiments}
	\subsection{Datasets}
	We conducted experiments using RAGTruth~\cite{RAGTruth} and Dolly-15k~\cite{dolly}. RAGTruth is a manually annotated RAG sample set generated by various LLMs. The Llama-7B part we used contains 510 hallucination samples and 479 normal samples, while the Llama-13B part contains 399 hallucination samples and 590 normal samples. Dolly-15k is a large model question-answering dataset covering multiple scenarios. We only used the closed question-answering scenarios oriented towards the RAG framework and filtered out samples with empty contexts. Consistent with~\citet{LRP4RAG}, GPT-4 is used to compare model outputs with standard answers to annotate the dataset.
	\subsection{Baselines}
	Detailed implementation of our method (SIRG) is provided in the Appendix \ref{sec:Implementation}. We compare SIRG with the following baselines: 
	\paragraph{Prompt}~\cite{RAGTruth} Through prompt engineering, we manually design LLM (Llama-7b and GPT-3.5-turbo) prompts to identify hallucinations.
	\paragraph{SelfCheckGPT}~\cite{SelfCheckGPT} SelfCheckGPT is employed to assess the consistency between sampled responses, calculating the probability of hallucination.
	\paragraph{Fine-tune}~\cite{RAGTruth} We fine-tune Llama-7b and Qwen-7b on the corresponding dataset to detect hallucinations.
	\paragraph{EigenScore}~\cite{Inside} This method utilizes the eigenvalues of the response covariance matrix to measure semantic consistency in the embedding space.
	\paragraph{SEP}~\cite{Semantic_entropy_probes} A linear probe trained on the hidden states of LLMs is utilized to detect hallucinations.
	\paragraph{LRP4RAG}~\cite{LRP4RAG} This is a method based on LRP, which directly feeds the contribution scores to SVM classifiers or LLMs for hallucination detection.
	\subsection{Main Results}
	We employ 3 evaluation metrics to compare SIRG with 9 state-of-the-art baselines on RAGTruth and Dolly-15k, with some results directly sourced from \citet{LRP4RAG}.
	
	Table \ref{llama} shows the comparison results on RAGTruth. 
	SIRG has strong performance across all metrics, even outperforming high-resource-consumption methods such as LLMs fine-tuning. 
	On \text{$\text{RAGTruth}_\text{Llama-7B}$} and \text{$\text{RAGTruth}_\text{Llama-13B}$}, SIRG ranks first with the highest F1, achieving improvements of $3.07\%$ and $5.78\%$ over the currently most advanced methods respectively.
	Prompt-based and self-validation-based methods (SelfCheckGPT) rely on pre-trained LLMs by designing prompts to enable single or multiple rounds of self-correction, which makes them highly unstable. Switching LLMs or prompts can significantly impact downstream tasks. For example, using different prompts with the same gpt-3.5-turbo resulted in a $44.89\%$ difference in recall. 
	Although gpt-3.5-turbo achieves $92.54\%$ recall in the SelfCheckGPT framework, its precision drops to $53.27\%$, indicating that LLMs blindly classify most samples as correct. This demonstrates that relying solely on pre-training knowledge is insufficient for accurate hallucination detection.
	The Fintune approach trains LLMs using specific RAGTruth data samples, yet its average performance remains only $62.74\%$ and $36.75\%$. We attribute this to the insufficient training data scale, which may inadvertently disrupt the general knowledge acquired through fine-tuning. Consequently, the fine-tuning outcomes are inferior to those of direct Prompt-based methods.
	Both EigenScore and SEP are methods based on vector space discriminators that lack direct contextual semantic information, making it difficult to adequately identify hallucination.
	LRP4RAG employs token-level aggregation of attribution vectors generated by the LRP algorithm to derive contextual relevance for model responses, representing a coarse-grained approach. This method introduces excessive noise of linking text, resulting in suboptimal performance during classifier training or discriminative tasks using LLMs.
	Our approach also employs the LRP algorithm, but it enhances the processing of substantive information in response texts by filtering out semantic noise and formally modeling it as a reasoning graph. This facilitates easier training of the downstream discriminator while helping humans understand the decision-making process of LLMs. See Fig.\ref{fig:hallu_frag} in Appendix \ref{sec:glue} for details of the example.
	
	\begin{table}[t]
		\resizebox{\linewidth}{!}{
			\begin{tabular}{llll}
				\toprule[2pt]
				\textbf{Model} & Precision & Recall & F1 \\
				\midrule[1pt]
				\multicolumn{4}{c}{\textbf{$\text{RAGTruth}_\text{Llama-7B}$}}\\
				\midrule[1pt]
				\text{$\text{Prompt}_\text{llama-7b}$} 				& $52.64\% $& $76.08\%$& $62.23\%$\\
				\text{$\text{Prompt}_\text{gpt-3.5-turbo}$} 		& $56.91\%$& $47.65\%$& $51.87\%$\\
				\text{$\text{SelfCheckGPT}_\text{llama-7b}$} 		& $53.32\%$& $83.53\%$& $65.09\%$\\
				\text{$\text{SelfCheckGPT}_\text{gpt-3.5-turbo}$} 	& $53.27\%$& \boldmath$92.54\%$& $67.62\%$\\
				\text{$\text{Fintune}_\text{llama-7b}$} 			& $62.50\%$& $65.75\%$& $63.58\%$\\
				\text{$\text{Fintune}_\text{qwen2-7b}$} 			& $61.76\%$& $64.34\%$& $61.90\%$\\
				EigenScore 											& $-$& $74.69\%$& $66.82\%$\\
				SEP 												& $-$& $74.77\%$& $66.27\%$\\
				\text{$\text{LRP4RAG}_\text{LLM}$} 					& $71.18\%$& $75.78\%$& $73.54\%$\\
				\midrule[1pt]
				SIRG (Ours) & \boldmath$73.64\%$& $79.83\%$& \boldmath$76.61\%$\\
				\midrule[1pt]
				\multicolumn{4}{c}{\textbf{$\text{RAGTruth}_\text{Llama-13B}$}}\\
				\midrule[1pt]
				\text{$\text{Prompt}_\text{llama-7b}$} 				& $41.02\% $& $ 56.64\%$& $47.58\%$\\
				\text{$\text{Prompt}_\text{gpt-3.5-turbo}$} 		& $47.58\%$& $44.36\%$& $45.91\%$\\
				\text{$\text{SelfCheckGPT}_\text{llama-7b}$} 		& $43.66\%$& $75.94\%$& $55.44\%$\\
				\text{$\text{SelfCheckGPT}_\text{gpt-3.5-turbo}$} 	& $43.01\%$& \boldmath$89.47\%$& $58.10\%$\\
				\text{$\text{Fintune}_\text{llama-7b}$} 			& $62.50\%$& $27.92\%$& $37.62\%$\\
				\text{$\text{Fintune}_\text{qwen2-7b}$} 			& $63.55\%$& $25.93\%$& $35.89\%$\\
				EigenScore 											& $-$& $67.15\%$& $66.37\%$\\
				SEP 												& $-$& $65.80\%$& $71.59\%$\\
				\text{$\text{LRP4RAG}_\text{LLM}$} 					& $77.14\%$& $74.58\%$& $75.86\%$\\
				\midrule[1pt]
				SIRG (Ours) & \boldmath$78.48\%$& $85.51\%$& \boldmath$81.84\%$\\
				\bottomrule[2pt]
			\end{tabular}
		}
		\caption{\label{llama}
			Overall precision, recall, and F1-score on RAGTruth with Llama-7B and Llama-13B.
		}
	\end{table}
	
	Table \ref{qwen} presents the comparative results of the Dolly-15k dataset. For the threshold-based benchmark model, we provide its optimal threshold parameters.
	Since the content generated by the LLM each time is random, fine-tuning to fit fixed responses is pointless in this scenario.
	On \text{$\text{Dolly}_\text{Qwen2.5-3B}$} and \text{$\text{Dolly}_\text{Qwen2.5-7B}$}, SIRG achieved F1 scores of $82.17\%$ and $89.17\%$ respectively, surpassing the state-of-the-art method LRP4RAG. 
	Due to the more unstable response of Qwen2.5-3B compared to Qwen2.5-7B, SIRG performs better on Qwen2.5-7B than on Qwen2.5-3B.
	
	\begin{table}[t]
		\resizebox{\linewidth}{!}{
			\begin{tabular}{llll}
				\toprule[2pt]
				\textbf{Model} & Precision & Recall & F1 \\			
				\midrule[1pt]
				\multicolumn{4}{c}{\textbf{$\text{Dolly}_\text{Qwen2.5-3B}$}}\\
				\midrule[1pt]
				Prompt												& $58.41\%$& $24.98\%$& $  34.99\%$\\
				SelfCheckGPT										& $ 67.32\%$& $32.47\%$& $43.88\%$\\
				EigenScore 											& $ 68.88\%$& $64.58\%$& $66.66\%$\\
				SEP 												& $77.94\%$& $ 79.19\%$& $ 78.56\%$\\
				\text{$\text{LRP4RAG}_\text{LLM}$} 					& \boldmath$80.55\%$& $ 82.91\%$& $ 81.71\%$\\
				\midrule[1pt]
				SIRG (Ours) & $72.10\%$& \boldmath$95.49\%$& \boldmath$82.17\%$\\	
				\midrule[1pt]
				\multicolumn{4}{c}{\textbf{$\text{Dolly}_\text{Qwen2.5-7B}$}}\\
				\midrule[1pt]
				Prompt												& $ 61.46\%$& $47.23\%$& $ 53.36\%$\\
				SelfCheckGPT										& $ 67.32\%$& $ 32.47\%$& $43.88\%$\\
				EigenScore 											& $58.57\%$& $70.03\%$& $63.79\%$\\
				SEP 												& $76.36\%$& $ 77.59\%$& $76.97\%$\\
				\text{$\text{LRP4RAG}_\text{LLM}$} 					& $79.60\%$& $ 82.20\%$& $ 80.80\%$\\
				\midrule[1pt]
				SIRG (Ours) & \boldmath$84.21\%$& \boldmath$96.00\%$& \boldmath$89.71\%$\\
				\bottomrule[2pt]
			\end{tabular}
		}
		\caption{\label{qwen}
			Overall precision, recall, and F1-score on Dolly-15k with Qwen2.5-3B and Qwen2.5-7B.
		}
	\end{table}

	\subsection{Faithfulness of LRP-based Internal Reasoning Graph}
	To verify the faithfulness of the internal inference graph constructed by SIRG, the same perturbation tests as~\cite{bakish2025revisiting} are employed (detailed in Appendix \ref{sec:perturbation}). 
	
\begin{figure}[t]
		\centering
		\includegraphics[height=0.48\textwidth,width=0.48\textwidth]{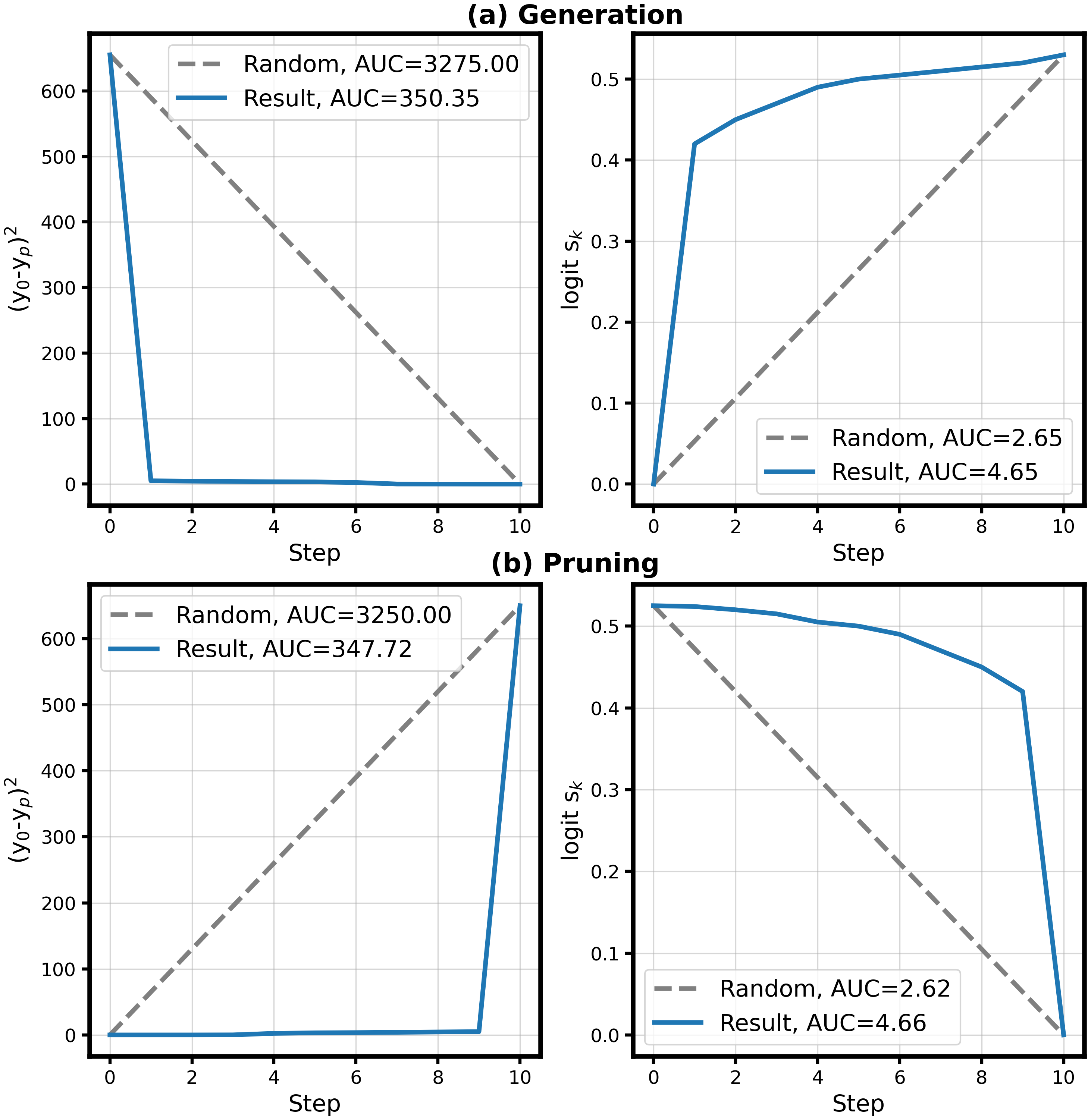}
		\caption{Perturbation tests on RAGTruth with Llama-7B. $(y_0-y_p)^2$ indicates the change of final embeddings before and after perturbation, while $logit s_k$ represents the average probability of the target semantic fragment. The dashed line shows the curve state after random  addition or pruning. Perturbation tests are conducted on 100 samples and mean of the above indicators are took.}
		\label{fig:faithful}
	\end{figure}

	We implement token-level blocking of semantic fragments in ten sequential steps based on their relevance, with results presented in Fig. \ref{fig:faithful}.
	When adding the semantic fragment deemed most relevant by LRP, the most significant changes are observed in the decrease of $(y_0-y_p)^2$ and the increase of $loght s_k$, indicating that this semantic fragment plays a pivotal role in LLM's computational process.
	After pruning semantic fragments based on relevance scores, those with lower relevance to the target exhibit negligible impact on both $(y_0-y_p)^2$ and $loght s_k$, whereas highly relevant fragments demonstrate substantial effects post-pruning. 
	This shows that our algorithm can effectively identify the source semantic fragment, which has a significant impact on the target semantic fragment.
	If the importance of contributions is randomly assigned, the result curve should change gradually with addition or pruning.
	Compared with the standard curve of random addition or pruning, our method demonstrates a significant $AUC$ advantage. 
	For quantitative comparison of this curve, please refer to the work of \citet{bakish2025revisiting}.
	
	\subsection{The Impact of Hyperparameters}
	
\begin{figure}[t]
		\centering
		\includegraphics[height=0.24\textwidth,width=0.48\textwidth]{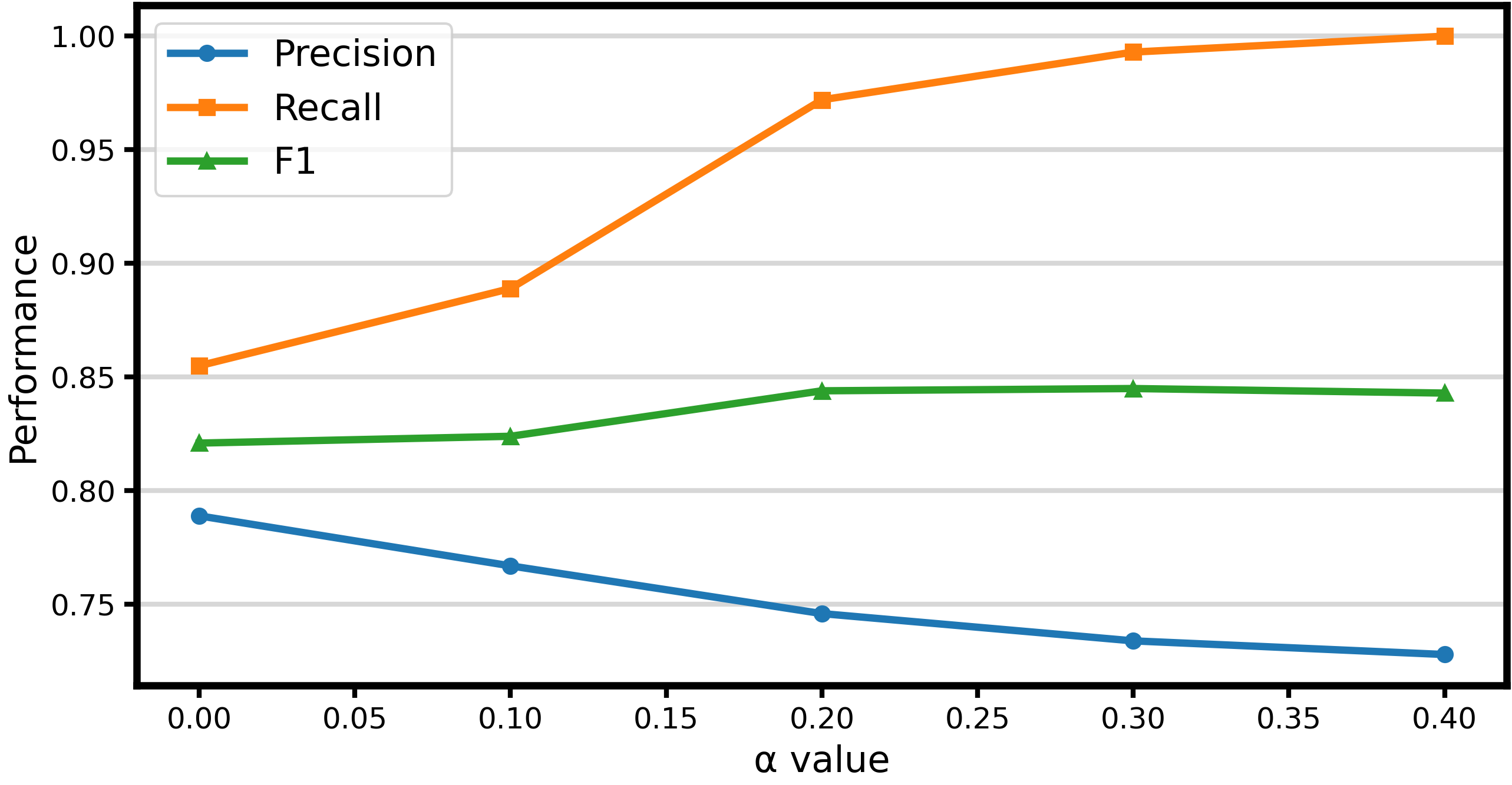}
		\caption{Overall precision, recall, and F1 score of Llama-13B on RAGTruth are evaluated by setting different $\alpha$ values.}
		\label{fig:alpha}
	\end{figure}

	For the classifier of SIRG, we focus on exploring the impact of $\alpha$ in Equation  \ref{eq:class}. 
	When $\alpha=0$, as long as there is one hallucinated semantic fragment, this response will be judged as a hallucinated sample, representing the strictest hallucination detection strategy. 
	As shown in Fig. \ref{fig:alpha}, at this point, SIRG has a relatively low recall for correct samples but a high precision in identifying hallucinations. 
	As the value of $\alpha$  increases, the detection strategy becomes increasingly lenient, so the pass rate for correct samples rises. 
	When $\alpha=0.4$, the recall for correct samples reaches $100\%$, but the precision decreases to $72.86\%$. Although adjusting $\alpha$ greatly impacts recall and precision, it has a relatively weak effect on the F1 score. 
	In different scenarios, we can dynamically adjust $\alpha$  based on the desired pass rate for correct samples.
	
	For the construction of the internal reasoning graph, we regulate the number of source semantic fragments by setting different $Topk$ values. 
	The performance of discriminators trained using graphs generated by various construction strategies is demonstrated in Table \ref{top}.
	As the number of source semantic fragments increases, the discriminator can obtain more semantic information. Consequently, the discriminators' F1 score progressively improves during the initial training phase, reaching its peak at $k=15$.
	An increase in $k$ will introduce edge contribution noise into the target semantic fragment, meaning many low-contribution semantic fragments are mistakenly fed to the discriminator for training.
	The discriminator may misinterpret the conflict between insignificant semantic fragments and target semantic fragments as hallucination, resulting in inferior performance at $k=20$ compared to $k=15$.
	For the adaptive discrete gradient strategy, it tends to only select 1 or 2 source semantic fragments. 
	Without sufficient information, the discriminator's performance is limited.
	\begin{table}[t]
		\resizebox{\linewidth}{!}{
			\begin{tabular}{lcccccc}
				\toprule[2pt]
				Top & 1 & 5 & 10 &15 &20 &Ada \\			
				\midrule[1pt]
				Precision								& $79.35\%$& $82.46\%$& $85.71\%$& $83.12\%$& $82.27\%$& $78.98\%$\\
				Recall									& $84.82\%$& $87.58\%$& $91.03\%$& $91.72\%$& $89.65\%$& $85.51\%$\\
				F1 										& $82.00\%$& $84.94\%$& $85.71\%$& $87.21\%$& $85.80\%$& $82.11\%$\\
				\bottomrule[2pt]
			\end{tabular}
		}
	\caption{\label{top}
			Overall precision, recall, and F1 score of Llama-13B on RAGTruth are evaluated by setting different $Top k$. Ada denotes the gradient-based adaptive construction strategy referred in Equation \ref{eq: grad}.
		}
	\end{table}
	
	\section{Conclusion}
	This paper first extends the token-level LRP algorithm to the semantic level within the autoregressive inference paradigm.
	Then we construct internal reasoning graphs using semantic fragments from RAG contexts and LLMs' responses, which faithfully model the dependencies of the internal reasoning process. 
	Based on that, we propose a framework, SIRG, for identifying faithfulness hallucinations in RAG. 
	SIRG achieves the performance of LLM-based detection frameworks using only a lightweight parameterized discriminator, demonstrating the effectiveness of our approach. 
	
	\section*{Limitations}
	For the internal reasoning graph construction of SIRG, since LRP requires computing internal model gradients for each token generation, this results in a high time complexity for the attribution score calculation phase.
	Future work will optimize LRP's computational objects from semantic fragment perspectives to reduce graph construction time. 
	We will also evaluate the semantic-level faithfulness of various attribution methods in capturing internal inference processes. 
	
	For the hallucination detection module of SIRG, although the linearization method has achieved great results, it is a naive way of using topological relation, which ignores the multi-hop dependency information and the subtle error propagation information in the internal reasoning graph.
	Future efforts will explore multi-angle applications of this graph, including adaptive node relationship aggregation via graph neural networks. 
	Additionally, developing low-resource hallucination discriminators remains a key research focus.
	
	\section*{Acknowledgments}
	We would like to appreciate anonymous reviewers for their valuable comments that help us to improve this manuscript.
	
	\section*{Ethics Statement}
	Based on the research presented in this paper, we acknowledge the ethical implications of developing hallucination detection methods for LLMs. While our work aims to enhance the reliability and trustworthiness of AI-generated content, we recognize that such techniques could potentially be misused to conceal model limitations or manipulate outputs in ways that undermine transparency. We affirm our commitment to responsible AI research by ensuring our method, SIRG, is designed to improve factual faithfulness rather than to deceive. All experiments were conducted using publicly available datasets with proper citations, and we have openly disclosed the limitations of our approach to avoid overstating its capabilities. We encourage the community to utilize this work for promoting accountability and interpretability in LLMs, and we emphasize the importance of continued ethical scrutiny as hallucination detection technologies evolve.
	
	\bibliography{custom}
	
	\appendix
	\section{Coarse-grained Processing}
	\label{sec:noise}
\begin{figure*}[t]
		\centering
		\begin{minipage}{0.49\textwidth}
			\centering
			\includegraphics[width=\linewidth]{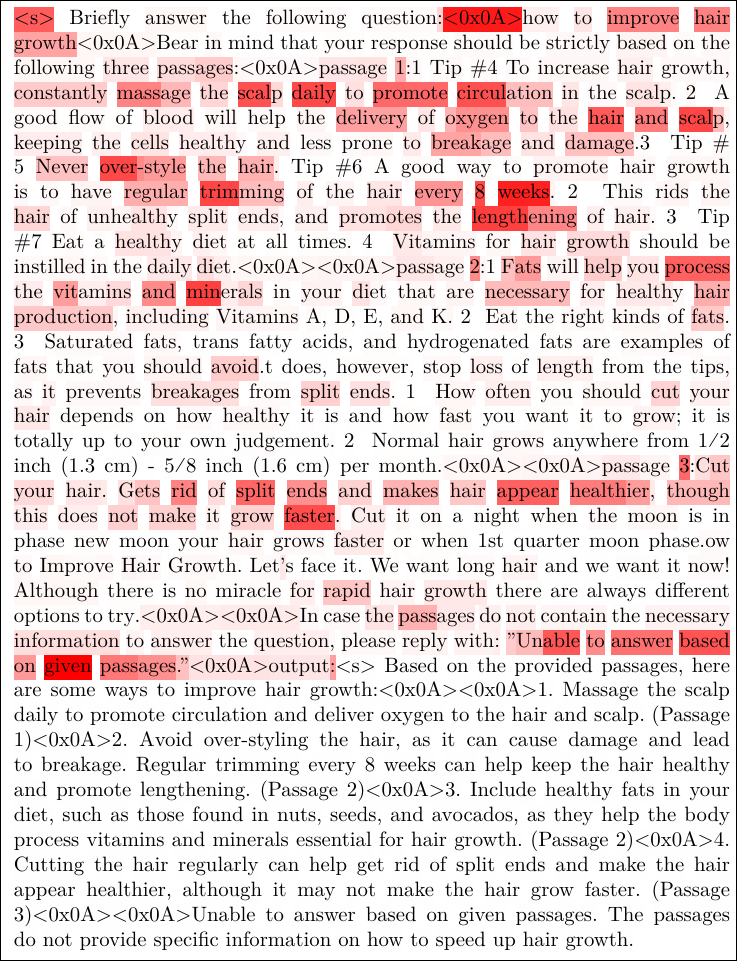}
		\end{minipage}
		\hfill
		\begin{minipage}{0.49\textwidth}
			\centering
			\includegraphics[width=\linewidth]{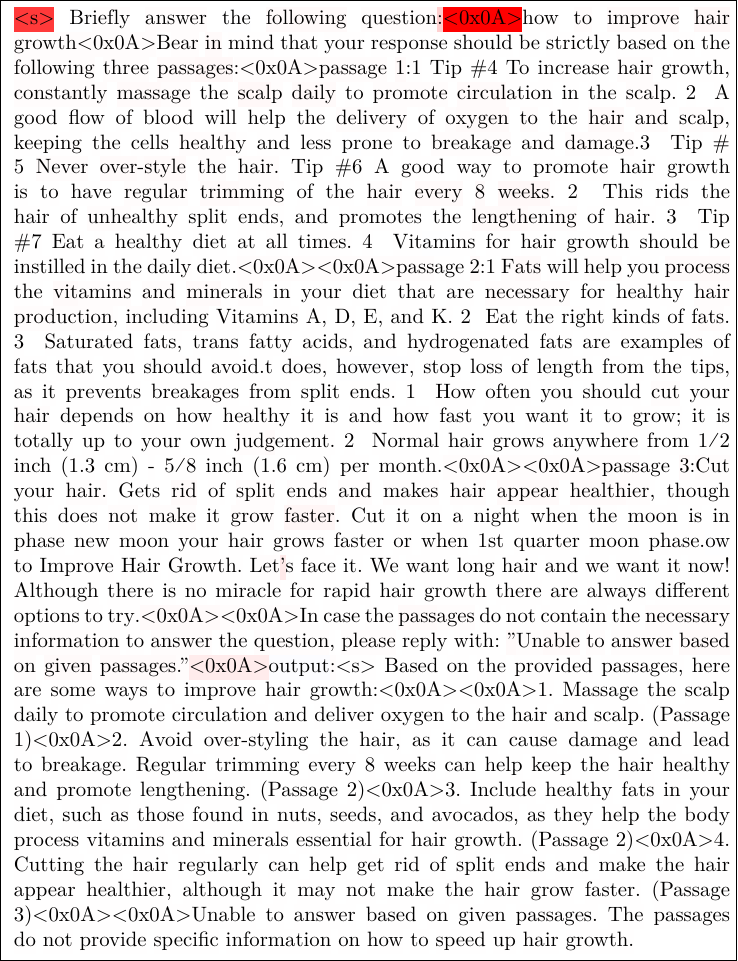}
		\end{minipage}
		\caption{The attribution of LLM responses is obtained by processing attribution features at a coarse granularity. The left figure shows the maximum value of the attribution vector for all tokens in the response, while the right figure shows the average value. The deeper the red color, the greater the token's contribution to the target response.}
		\label{pic:noise}
	\end{figure*}

	\citet{LRP4RAG} employs the maximum and average value of the LRP attribution vector for all tokens to get the attribution distributions of the whole response. As shown in Fig. \ref{pic:noise}, using the maximum value method accumulates the context that contributes the most to each response token, resulting in a noisy contribution distribution. Using the average value method dilutes the context with a high contribution to the substantive word with a large number of linking words, leading to a significantly lower contribution distribution.
	
	\section{Linking Tokens and Substantive Tokens}
	\label{sec:glue}
	In the sentence ``The 15 minute guideline starts from the point at which you put your scones in a preheated oven'', ``15 minute'', ``preheated oven'', etc., contain rich semantic information and are considered as substantive tokens, while ``The'', ``which'', etc., serve to enhance sentence fluency in the document and are therefore considered linking tokens. 
	
	The above description of the linking and substantive token is identified from a human perspective. 
	From an LLM's standpoint, due to semantic drift and dataset bias during training, the linking and substantive tokens inferred by the LLM may differ from human perceptions. 
	When LLMs generate tokens by treating human-considered substantive content as linking, it results in the faithfulness hallucination.
	The cognitive gap between humans and LLMs makes this phenomenon hard to detect directly.
	
\begin{figure*}[t]
		\centering
		\begin{minipage}{0.325\textwidth}
			\centering
			\includegraphics[width=\linewidth]{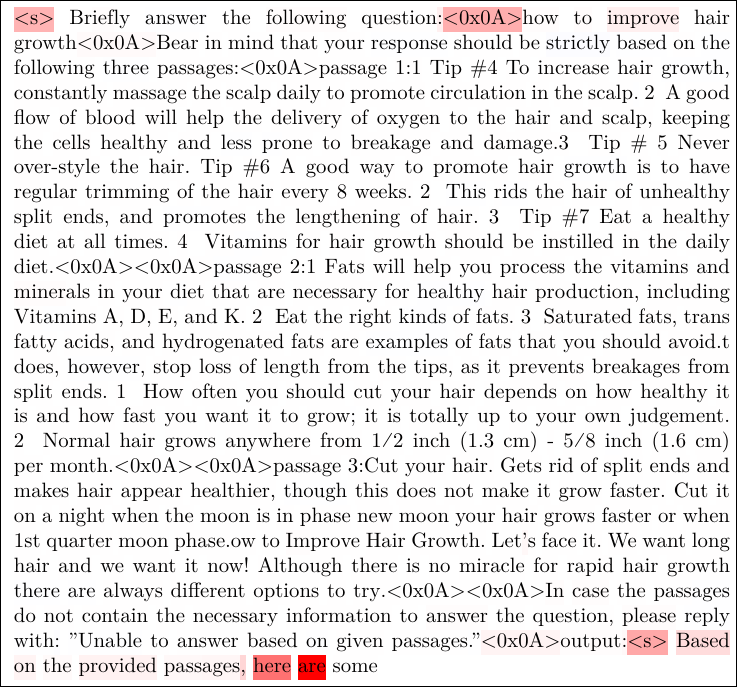}
		\end{minipage}
		\hfill
		\begin{minipage}{0.325\textwidth}
			\centering
			\includegraphics[width=\linewidth]{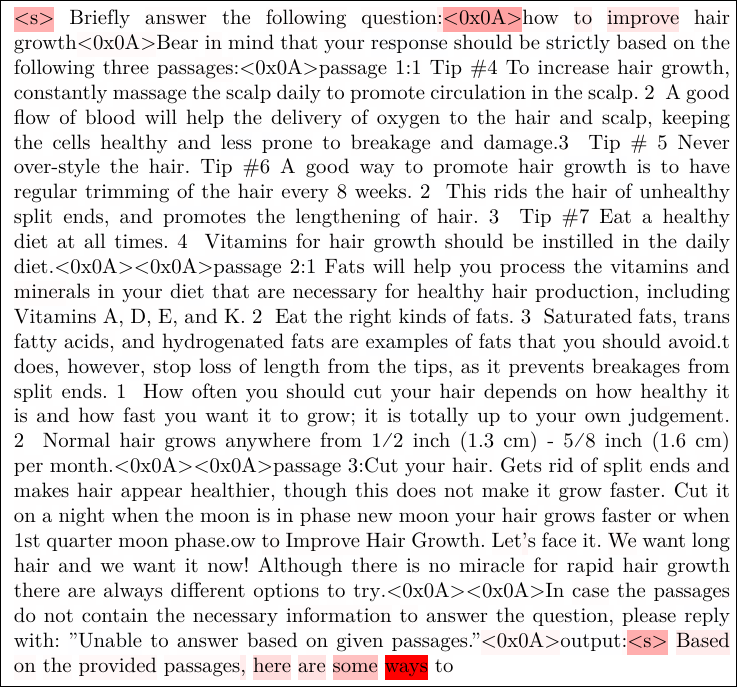}
		\end{minipage}
				\hfill
		\begin{minipage}{0.325\textwidth}
			\centering
			\includegraphics[width=\linewidth]{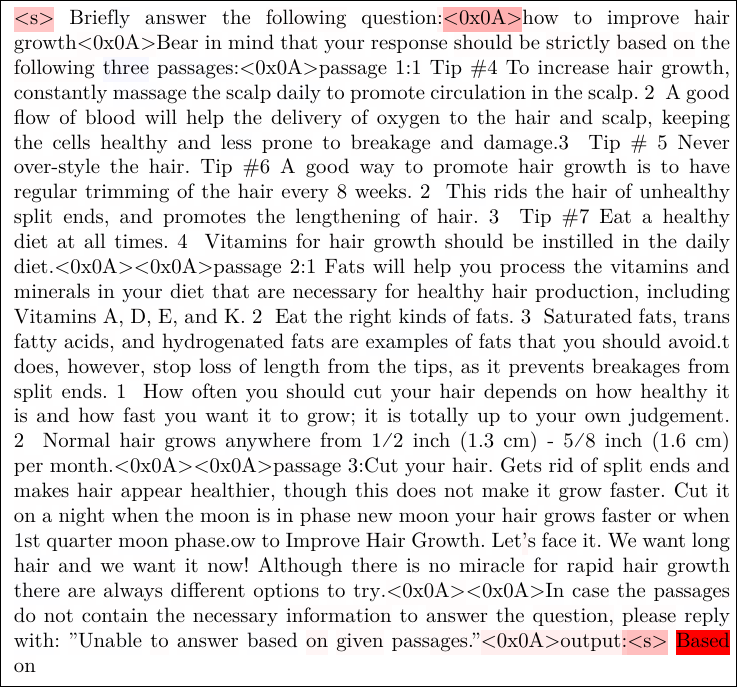}
		\end{minipage}
				\hfill
		\begin{minipage}{0.325\textwidth}
			\centering
			\includegraphics[width=\linewidth]{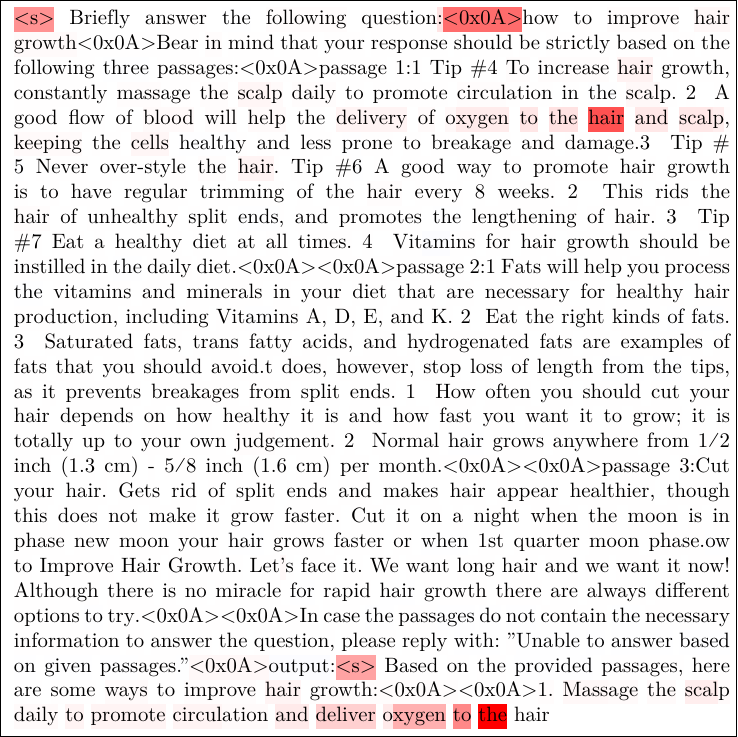}
		\end{minipage}
				\hfill
		\begin{minipage}{0.325\textwidth}
			\centering
			\includegraphics[width=\linewidth]{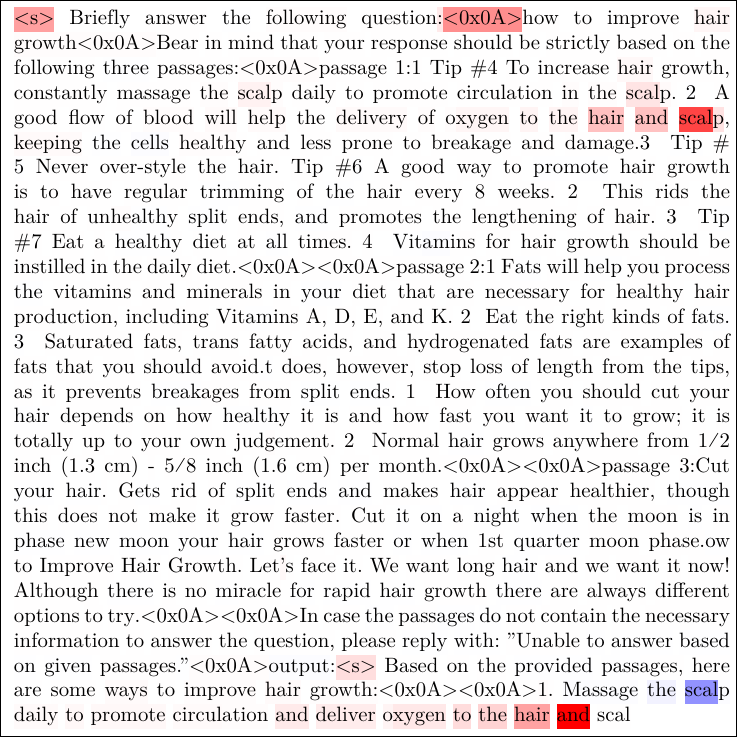}
		\end{minipage}
				\hfill
		\begin{minipage}{0.325\textwidth}
			\centering
			\includegraphics[width=\linewidth]{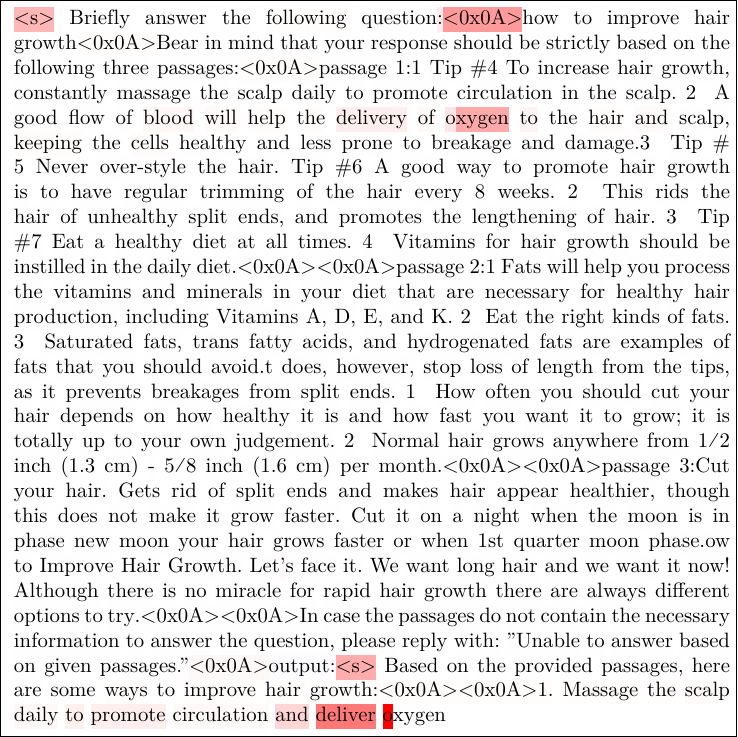}
		\end{minipage}
		\caption{Contribution scores of the next generated token. The upper three boxes show the distribution of attribution scores for the next generated linking token, while the lower three boxes display the substantive token. The deeper the red color, the greater the token's contribution to the last token.}
		\label{pic:glue}
	\end{figure*}

\begin{figure}[ht]
		\centering
		\begin{minipage}{0.48\textwidth}
			\centering
			\includegraphics[width=\linewidth]{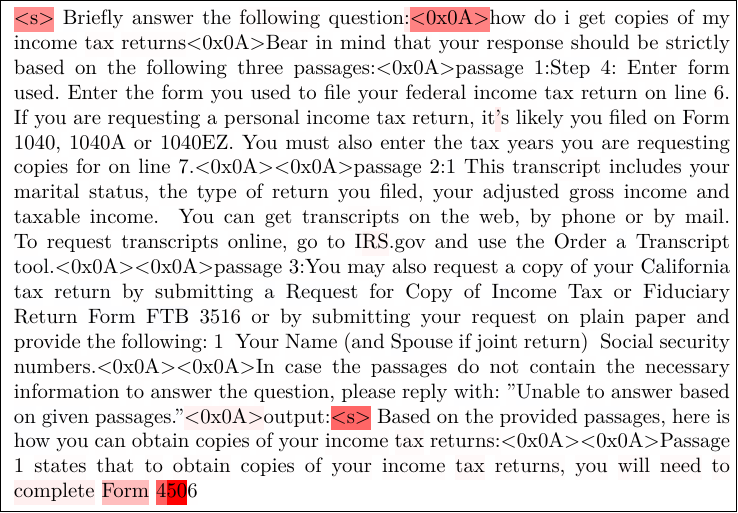}
		\end{minipage}
		\hfill
		\begin{minipage}{0.48\textwidth}
			\centering
			\includegraphics[width=\linewidth]{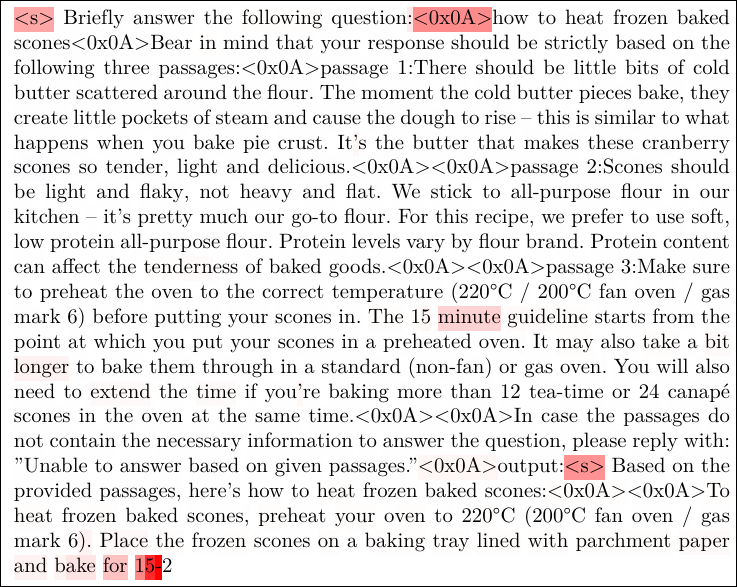}
		\end{minipage}
		\caption{The contribution score of the next generated token (labeled as hallucination). The deeper the red color, the greater the token's contribution to the last token.}
		\label{pic:hallu}
	\end{figure}

	We employ the LRP to perform token-level attribution of the next token generated by the LLM, and visualize the results using a heat map.
	As shown in Fig.\ref{pic:glue}, linking tokens rely more on words generated earlier within the same sentence, whereas substantive tokens also depend on words in the long-distance context.
	When the model generates hallucination tokens, the heat map in Fig.\ref{pic:hallu} reveals that their attribution distribution exhibits characteristics of linking tokens. 
	From a human perspective, however, these tokens should be substantive tokens in the RAG system that strictly depend on contextual content. 
	In other words, LLMs process substantive tokens as linking, resulting in faithfulness hallucination.
	
\begin{figure}[t]
		\centering
		\includegraphics[width=0.48\textwidth]{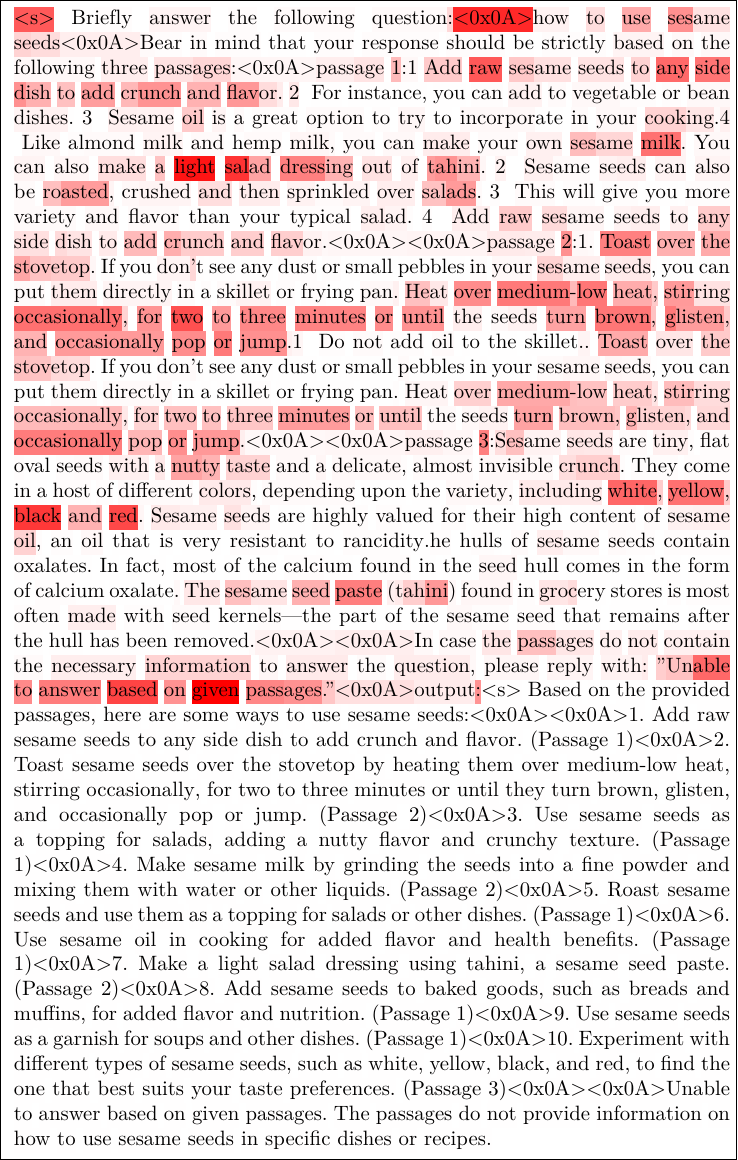}
		\caption{The superimposed token-level attribution distribution of the response (start with ``<s> Based on''). The deeper the red color, the greater the token's contribution to the last token. Based solely on the overlapping attribution distribution, it is difficult to determine whether the response contains hallucination fragments.}
		\label{fig:max_fragment}
	\end{figure}

\begin{figure}[t]
		\centering
		\includegraphics[width=0.48\textwidth]{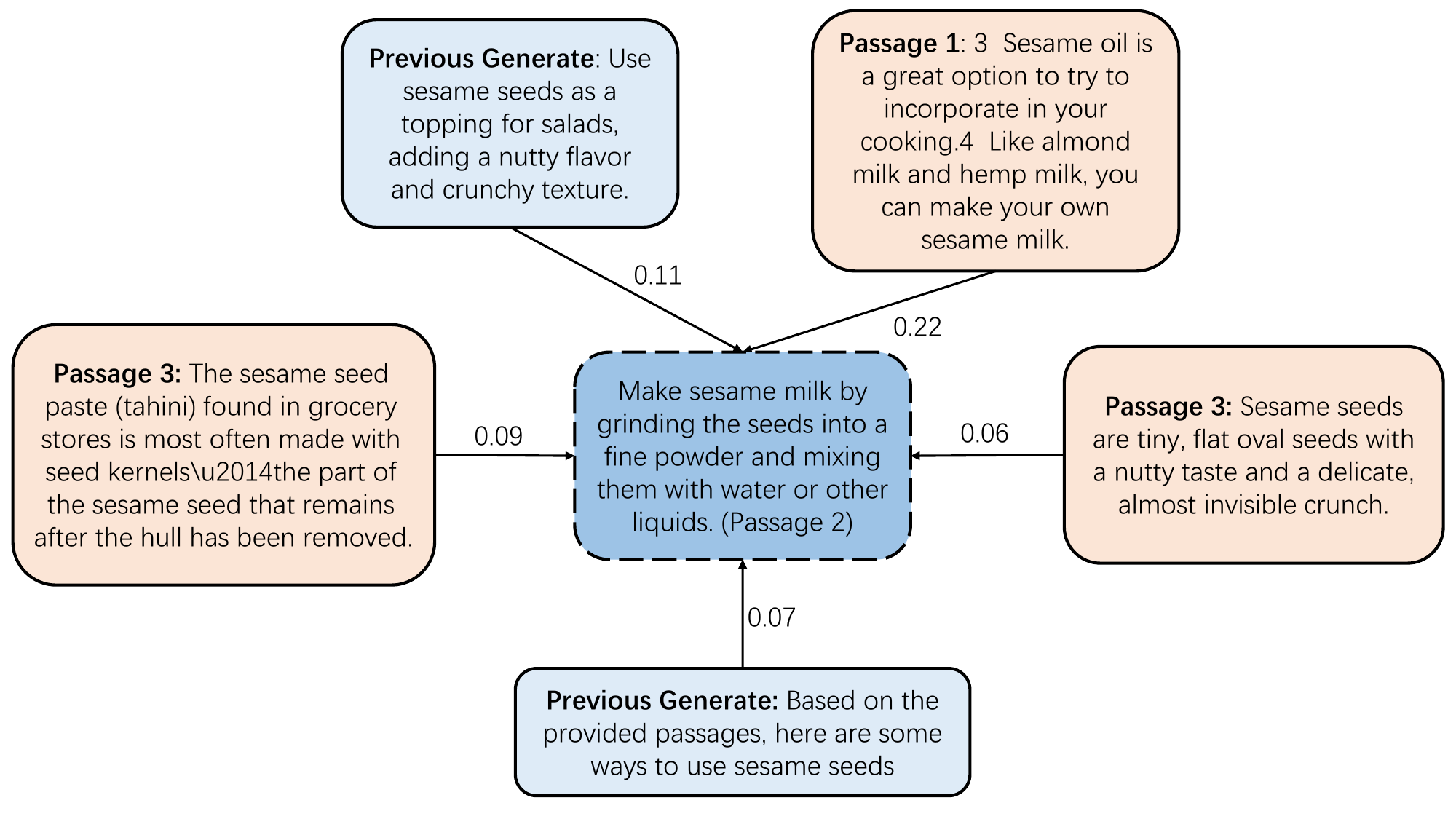}
		\caption{The internal reasoning graph constructed based on the target fragment in the response shown in Fig. \ref{fig:max_fragment}. The target fragment clearly originates from passage 1, passage 3 and previous generation, contradicting the annotation that it comes from passage 2. Thus, the fragment within the dashed box is most likely a case of faithfulness hallucination.}
		\label{fig:hallu_frag}
	\end{figure}

	Faithfulness hallucination is a semantic phenomenon where LLM responses contain content that is semantically inconsistent or contextually absent. 
	When isolated from context, it is challenging to determine whether LLMs exhibit faithfulness hallucination at the token level alone. 
	As shown in Fig.\ref{fig:max_fragment}, the sample's golden label is ``Have Hallucination'' and the reason is ``LOW INTRODUCTION OF NEW INFORMATION. Original: This might be correct, however, the exact way to make sesame milk is not directly stated in the passages. Generative: ...by grinding the seeds into a fine powder and mixing them with water or other liquids. (Passage 2)''. 
	However, direct analysis of the superimposed token-level attribution distribution makes it difficult to identify the hallucination fragment. 
	By constructing the target fragment as a semantic-level reasoning graph using our method (as shown in Fig.\ref{fig:hallu_frag}), we can easily identify that the information in the target fragment primarily originates from Passage 1, Passage 3, and the LLM's previous generation, indicating a high probability of the faithfulness hallucination.
	
	\section{LRP Rules}
	\label{sec:lrp}
	To obtain token-level correlations within LLM, we backpropagate output correlations through each transformer layer. 
	
	For the densely connected modules within the transformer architecture, we utilize the local Jacobian matrix $ J_{ji}=\frac{\partial z_j}{\partial x_i} $, which represents the gradient of the layer's output $z_j$ with respect to its input $x_i$, as the weight for assigning correlations. Specifically, for the multi-layer perceptrons within the model, they typically consist of a linear layer followed by an additional nonlinear function:
	\begin{equation}
		\begin{split}
			& z_j=\sum _i \bm{W_{ji}} x_i+b_j \\
			& a_j=\sigma (z_j)
		\end{split}
		\label{eq:line}
	\end{equation}
	where $W_{ji}$ represents the model's parameter and $\sigma$ denotes the nonlinear function. By linearizing the linear layer in Equation \ref{eq:line} at any point $x \in \mathbb{R}^N $, the basic LRP rule can be derived:
	\begin{equation}
		R_i^{l-1}=\sum_j \bm{W_{ji}} x_i \frac{R_j^l}{z_j (\bm{x})+ \varepsilon} 
	\end{equation}
	Since element-wise nonlinearity has only a single input variable and output variable, the decomposition of attribution results is the operation itself. Therefore, the total input correlation $R_i^l$ can only be allocated to a single input variable.
	\begin{equation}
		R_i^{l-1}=R_i^l
		\label{eq:eq}
	\end{equation}
	We apply the Equation \ref{eq:eq} to all element-wise operations of single-input and single-output variables.
	
	For the multi-head attention module in the transformer architecture, let $Q$, $K$, and $V$ represent the query, key, and value matrices, respectively. The scaled dot product attention calculates the attention weight $A$ and output value $O$ as follows:
	\begin{equation}
		\begin{split}
			& \bm{A}=softmax(\frac{\bm{Q} \cdot \bm{K}^T}{\sqrt{d_k}}) \\
			& \bm{O}=\bm{A} \cdot \bm{V} \\
			& softmax_j (\bm{x})=\frac{e^{x_j}}{\sum_k e^{x_k}}
		\end{split}
	\end{equation}
	Taylor expansion of the softmax function at reference point $x$ yields the following correlation propagation rules:
	\begin{equation}
		R_i^{l-1}=x_i (R_i^l - s_i \sum_j R_j^l )
	\end{equation}
	Among them, $s_i$ represents the i-th output of the softmax function. For the double matrix multiplication part in self-attention computation, since $f(0,0)=0$ holds, it is necessary to decompose the matrix multiplication into a form without bias terms. For this purpose, we decompose matrix multiplication into affine operations containing summation and bilinear parts containing element-wise multiplication.
	\begin{equation}
		\bm{O}_{jp}=\sum _i \bm{A}_{ji} \bm{V}_{ip}
	\end{equation}
	For the bilinear multiplication summation part mentioned above, we decompose it into element-level multiplication operations with $N$ input variables:
	\begin{equation}
		f_j (x)=\prod_i^N x_{ji} 
	\end{equation}
	The following uniform correlation propagation rules can be obtained at the reference point $x$ using the Shapley method (with a baseline of zero) or the Taylor decomposition method:
	\begin{equation}
		R_{ji}^{l-1}=\frac{1}{N} R_j^l
	\end{equation}
	Therefore, based on the correlation propagation calculation formulas of each sub-part in the self-attention module mentioned above, we can obtain the following correlation propagation rules in the self-attention layer:
	\begin{equation}
		R_{ji}^{l-1}=\sum_p \bm{A}_{ji} \bm{V}_{ip} \frac{R_{jp}^l} {2\bm{O}_{jp}+\varepsilon}
	\end{equation}
	There is no bias term for absorption correlation in this rule, and the amount of $\varepsilon$ absorption can be ignored. By adopting this rule, we strictly adhere to the conservation property in the process of correlation propagation.
	
	Regarding the correlation propagation rule of LayerNorm layer, when using common $\varepsilon=10^{-6}$ and $Ver[x]=1$, it actually absorbs $99\%$ of the correlation. 
	Therefore, linearizing at $x$ is meaningless. By using Taylor expansion to decompose LayerNorm or RMSNorm with reference point $0$, we can obtain the same identity association propagation rule as Equation \ref{eq:eq}.
	
	By using the correlation backpropagation based on the above rules, we can obtain the contribution score of the model input to the probability value output. 
	Intuitively, the higher the contribution score, the greater the impact of the input label on a specific output label. 
	By combining these contribution vectors into a matrix $\bm{R}_i$, we can obtain a correlation matrix that is faithful to the internal inference process of the model. 
	The proof is detailed in~\cite{AttnLRP}.
	
	\section{Implementation Details}
	\label{sec:Implementation}
	We employ LXT 2.0 to compute LRP attribution scores for both Llama and Qwen. For substantive word extraction, we utilize Spacy and Stanza to identify nouns, named entities, noun phrases, and negations in the text. During training, the AlignScore pre-trained Roberta model with 124M parameters is used. The batch size is set to $16$, and we perform $100$ iterations on the training set using the Adam optimizer at a learning rate of $1e-5$. The model with the best F1 score is selected for the final result.
	
	\section{Perturbation Tests}
	\label{sec:perturbation}
	The perturbation tests employed by \citet{bakish2025revisiting} are divided into two types of disturbance: generation and pruning.
	The generation process progressively incorporates semantic fragments starting from null, ordered by relevance from highest to lowest. 
	An approach that faithfully replicates the real decision-making process of LLMs will identify the most impactful semantic fragments. 
	When these fragments are added, the embedding distance of this model is significantly reduced compared to the previous one, and the logit (calculated against the predicted target semantic fragment) shows a marked increase.
	The pruning method starts by masking the semantic fragments with the lowest relevance and gradually progresses towards the semantic fragments with higher importance. 
	Removing low-impact semantic fragments should preserve model prediction stability.
	Following \citet{ali2022xai}, the final metrics were quantified using Area-Under-Curve (AUC), capturing model accuracy relative to the percentage of masked semantic fragments, from $0\%$ to $100\%$.

\end{document}